# Application of an automated machine learning-genetic algorithm (AutoML-GA) coupled with computational fluid dynamics simulations for rapid engine design optimization

Opeoluwa Owoyele[1*], Pinaki Pal[1], Alvaro Vidal Torreira[2], Daniel Probst[3], Matthew Shaxted[2], Michael Wilde[2], Peter Kelly Senecal[3]

[1]Argonne National Laboratory, Energy Systems Division, Lemont, IL
[2]Parallel Works Inc., Chicago, IL
[3]Convergent Science, Inc., Madison, WI

*Corresponding Author: oowoyele@anl.gov, phone: +1 919-412-9815.

## Abstract

In recent years, the use of machine learning-based surrogate models for computational fluid dynamics (CFD) simulations has emerged as a promising technique for reducing the computational cost associated with engine design optimization. However, such methods still suffer from drawbacks. One main disadvantage of is that the default machine learning (ML) hyperparameters are often severely suboptimal for a given problem. This has often been addressed by manually trying out different hyperparameter settings, but this solution is ineffective in a high-dimensional hyperparameter space. Besides this problem, the amount of data needed for training is also not known *a priori*. In response to these issues that need to be addressed, the present work describes and validates an automated active learning approach, AutoML-GA, for surrogate-based optimization of internal combustion engines. In this approach, a Bayesian optimization technique is used to find the best machine learning hyperparameters based on an initial dataset obtained from a small number of CFD simulations. Subsequently, a genetic algorithm is employed to locate the design optimum on the ML surrogate surface. In the vicinity of the design optimum, the solution is refined by repeatedly running CFD simulations at the projected optimum and adding the newly obtained data to the training dataset. It is demonstrated that AutoML-GA leads to a better optimum with a lower number of CFD simulations, compared to the use of default hyperparameters. The proposed framework offers the advantage of being a more hands-off approach that can be readily utilized by researchers and engineers in industry who do not have extensive machine learning expertise.

# 1      Introduction

Unwanted pollutant emissions from fossil-fueled vehicles in the transportation sector, partially contributing to changes in the global climate, have led to increased concerns in recent decades. Related to this are concerns regarding the reduction in air quality and the effects of pollutants on human health, causing governments and regulatory bodies to impose increasingly stringent emission standards on automakers.

In this context, simulation-based design optimization can play a crucial role in helping automakers improve fuel economy while meeting emission constraints. On one hand, using simulations allows for the effects of design variables (e.g., geometrical features), which are otherwise difficult to vary in experiments, to be studied. Simulations can enable engineers to perform more extensive exploration of the design space while saving costs associated with building multiple prototypes. On the other hand, however, simulation-driven design optimization (SDDO) faces a few challenges. First, the objective function to be optimized is usually a non-linear function of the design variables and their associated complex interactions. Most problematic, however, is the sheer amount of computing resources required to evaluate the objective function, which involves running multiple CFD simulations. Running full-scale engine CFD simulations require the use of several compute cores over a long period of time. Therefore, many optimization schemes historically used for engine design optimization (genetic algorithms, response surface methods)[1-3] suffer from slow convergence times and can take several months to find optimal designs.

In recent years, some studies have been performed to alleviate this bottleneck by reducing the number of CFD simulations required to obtain optimal design variables. One class of methods is sequential and operates by running simulations successively over design iterations while showing decreased runtime or number of design iterations required.[4-7] Other methods employ surrogate models, where all or most of the simulations are run *a priori*. Moiz et al.[8] performed the optimization of a compression-ignition internal combustion engine running on a gasoline-like fuel using a machine learning-genetic algorithm approach (ML-GA). The surrogate machine learning model was based on a Super Learner[9] approach, where multiple base learners were combined into a single surrogate model. The results of this study showed that the required number of CFD simulations and the overall runtime were reduced compared to a CFD-GA, with a significant source of savings coming from the potential to run all the simulations in parallel in the presence of sufficient computing resources. Badra et al.[10, 11] employed the ML-GA approach developed by Moiz et al.[8], but replaced the genetic algorithm with a grid gradient algorithm (GGA), which led to more repeatable global optima. Badra et al.[11] also investigated the effects of modest extrapolations to the design space and incorporated an active learning technique where the solution was refined by sequentially running more simulations around the global optimum. While these surrogate-based techniques have shown promise in solving the issues facing SDDO, a major disadvantage that hinders wide deployment of such methods relates to the selection of hyperparameters for the machine learning models. In previous studies, hyperparameters have been selected using default values combined with manual tuning, while using domain knowledge in machine learning as a guide. However, reducing the need for prior machine learning when using such optimization approaches is important, if such technologies are to be adopted widely by the industry.

In a previous study, Owoyele et al.[12] introduced an automated machine learning-genetic algorithm technique (AutoML-GA), which improves upon ML-GA by incorporating automated hyperparameter tuning along with an active learning loop to improve performance. In this study, the process of selecting hyperparameters manually was replaced by a Bayesian approach,[13, 14] and it was shown that the hyperparameters selected in this fashion led to more accurate surrogate models and a shorter active learning loop. In the previous study, however, the ground-truth objective function was assumed to be a surrogate model from the Moiz et al.[8] study. In this work, the authors build on the previous AutoML-GA study by coupling AutoML-GA directly with 3D CFD simulations. The paper is organized as follows. First, a brief overview of the base learners and AutoML-GA optimization approaches is presented. Thereafter, the accuracy of the ML surrogates obtained using optimally selected hyperparameters is evaluated and compared to the errors that arise from using default hyperparameters. Besides, differences between the number of simulations required to reach the design optimum are compared for both approaches. Criteria for assessing convergence and terminating the active learning loop are also introduced. The paper ends with some concluding remarks and thoughts for potential future directions.

## 2 Automated ML-GA

### 2.1 ML-GA

As discussed in the previous section, one promising approach that has been used to accelerate SDDO of IC engines is the use of machine learning surrogates. Here, the goal is to replace a CFD simulation with a surrogate model that has a much faster execution time. Typically, this process involves running a large number of CFD simulations and then creating a database that contains the design variables and the corresponding objective values as determined by the CFD simulations. Afterward, a machine learning model is trained to learn from this database and thereby provide an approximation to the true objective as obtained from CFD. This trained machine learning model, if accurate enough, can be used with a conventional global optimizer (e.g., a genetic algorithm) in place of CFD. The use of surrogate models can provide savings in runtime because the simulations (to create the database) can be run in parallel and independent of each other. This is in contrast to directly coupling a genetic algorithm with CFD, where the design points sampled for a given iteration depend on the previous iterations.

A key component in the success of surrogate-based approaches, unsurprisingly, lies in the accuracy of the surrogate model used. In the ideal case where the differences between the CFD-predicted objective and the surrogate model's predictions are infinitesimal, the design optimum obtained from the optimization would be the same. However, in practice, errors are inevitable and thus differences between the optimum obtained from CFD and the projected optimum from the surrogate model are present. Overall, it is expected that more accurate surrogates will provide better objective values when the results are validated against CFD. When the surrogate model does not closely match the CFD-derived objective, optima obtained using the surrogate models may be far from the true optimum that would be obtained from CFD. One technique that has been shown to improve the prediction accuracy of machine learning models combines individual learners for prediction as opposed to using only one. This method is called a Super Learner approach. The individual machine learning models that make up the Super Learner are referred to as the base learners. The surrogate models employed in this study utilize this approach, employing the base learners described below:

Regularized polynomial regression (RPR): This involves the transformation of the original set of design variables into a new set using a polynomial transformation. Linear regression with an L2 norm regularization uses the obtained features to predict the quantity of interest. The purpose of using regularization is to prevent oscillations at the boundaries of the design space, while the polynomial transformation allows for non-linear mappings.

Support vector regression (SVR): SVR[15] learns a function by finding the line that minimizes the cost, based on a margin defined by a tolerance. Non-linear functions can be learned through the use of kernels. This work employs a $\upsilon$-SVR approach, where the tolerance is calculated and the ratio of the number of support vectors to data samples is set by the user.

Kernel ridge regression (KRR): KRR is similar to SVR but minimizes the squared error between the actual and predicted values. It uses a kernel transformation to capture non-linear functions and L2 regression to mitigate overfitting.

Extreme gradient boosting (XGB): XGB[16] is a tree-based algorithm, comprised of several regression trees, which are trained by recursively bifurcating nodes and setting the predictions to constant values in each obtained partitions. These trees, which are weak learners when operating individually, are combined to form a strong learner. Extreme gradient boosting also features regularization techniques to improve generalization and fast inference times.

Artificial neural networks (ANNs): ANNs are machine learning models that mimic biological neural networks. They operate by applying matrix operations and a non-linear or piecewise linear activation function over successive hidden layers.

All the learners used in this work were developed in Python using packages such Numpy[17] and Scikit-learn.[18] The individual base learners are first trained on the dataset. Afterward, the overall prediction from the Super Learner, $\Phi$, given $M$ base learners is obtained by:

$$\Phi = \sum_{i=1}^{M} \varphi_i W_i \tag{1}$$

In Eq. 1, $\varphi_i$ and $W_i$ refer to the prediction and weight of the $i$th base learner, respectively. $W$ is obtained by using a non-negative least squares method to find the weights that provide the smallest prediction errors. In the event that a base learner $i$ would worsen the accuracy of the Super Learner, the solution for $W_i$ would be zero, ensuring that the base learner would have no contribution to the Super Learner. In this way, this approach guarantees that the error would be smaller or equal to that of the best base learner. Overall, the goal of using several base learners is to compensate for errors from the individual base learners. This Super Learner can serve as a surrogate for CFD, with a genetic algorithm is used to find the global optimum.

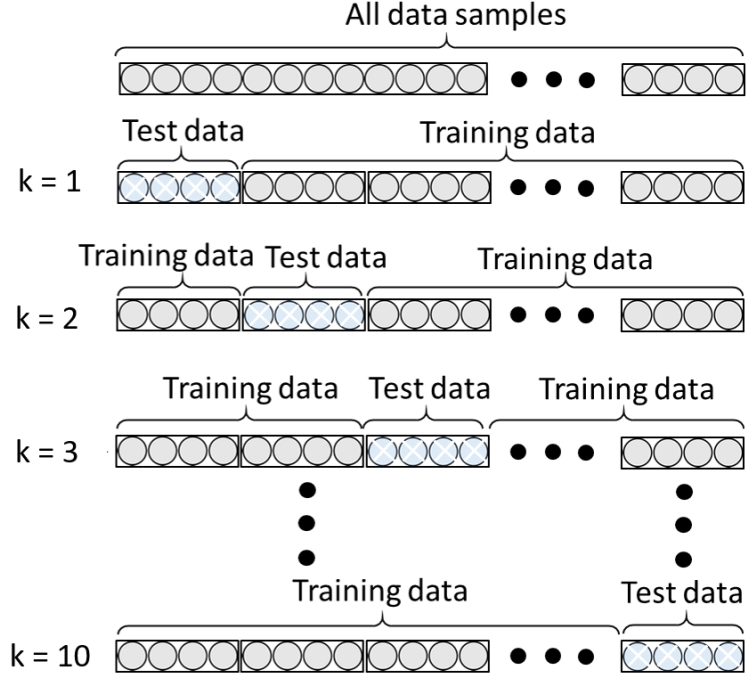

Figure 1. An illustration of the *k*-fold validation approach used in this study.

One common way to avoid overfitting when training machine learning models is to set aside a portion of the data (known as the test data) that is not used for training. The performance of the machine learning model on the test data, as opposed to the training data, provides an assessment of the model performance during deployment. In this case, data generation is an expensive process, involving three-dimensional (3D) multi-physics CFD simulations. So, setting aside a portion of the data for testing the model would be a waste of scarce training data. Therefore, a *k*-fold validation process is utilized to assess the models. It involves dividing the entire data, consisting of *N* samples, into *10* equal test partitions of size *N/10*. The training process is repeated 10 times, each time using a different test partition as the test data and the remaining *9N/10* samples as the training data. After all the test partitions have been used up, the mean squared errors (MSE) obtained from the test partitions are averaged to find overall model performance. For the remainder of this paper, the mean squared errors obtained in this manner are referred to as the cross-validated mean squared errors (CVMSE). The *k*-fold validation process used in this study is illustrated in Fig. 1. This *k*-fold validation process was used to calculate the optimal weights of the base learners in the Super Learner and for calculating the loss function when performing automated hyperparameter selection (as described in Section 2.2).

In general, it is expected that for very small datasets, the CVMSE may be much worse than the training MSE, since the test data may not be statistically representative of the full dataset. This is the case in most CFD-driven design optimization endeavors since the generation of large amounts of data is typically limited by computational resources. However, using small datasets with high CVMSE, as discussed above, leads to large discrepancies between the actual global optimum and the optimum obtained from the surrogate-assisted optimization. In many cases, the surrogate model predicts a very favorable objective value that ends up being underwhelming when evaluated with CFD. In such cases, the CVMSE is expected to improve as we include more data. Nevertheless, by optimally setting the hyperparameters of the machine learning models for the dataset of interest, the prediction accuracy can be improved without the need for generating additional expensive CFD data samples.

## 2.2 Automated hyperparameter selection and active learning

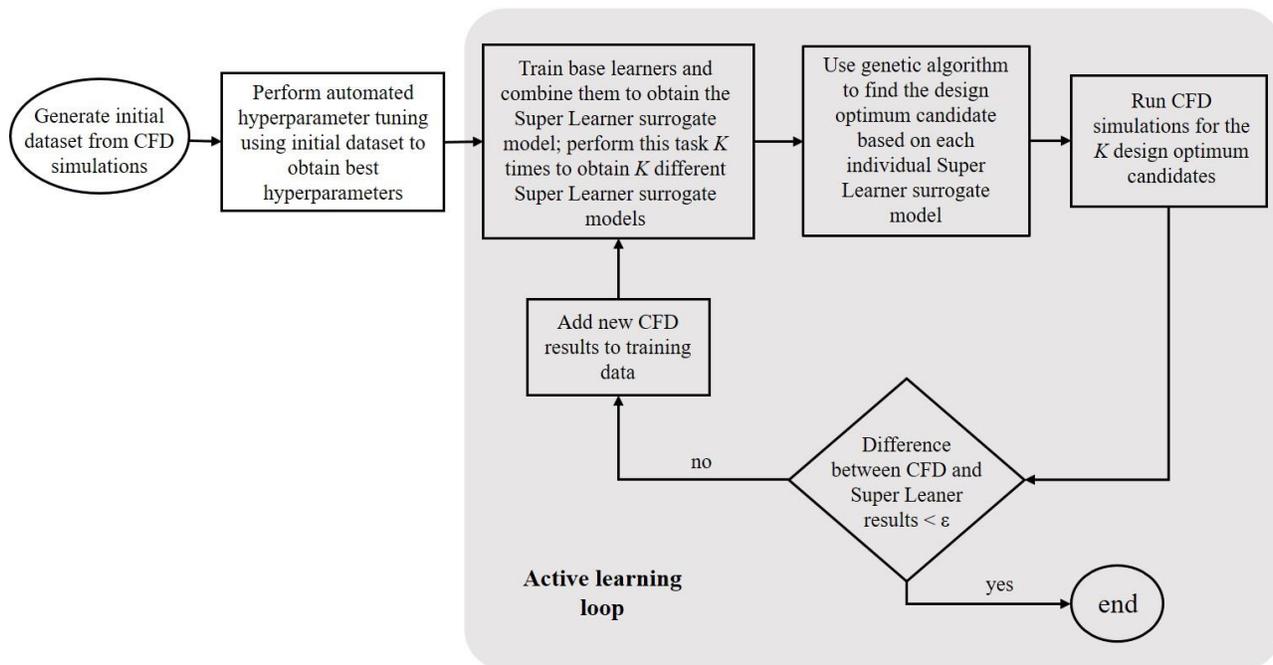

Figure 2. A schematic describing AutoML-GA. In this study, $K = 5$ and a maximum of 20 simulation batches (active learning loops), including the initial dataset, are run.

[Insert Figure 2]

The automated hyperparameter selection and active learning process, illustrated in Fig. 2, will be discussed in this section. Machine learning is essentially an optimization problem, where the goal is to find the model parameters that minimize a loss function of interest. For instance, while training neural networks for a regression problem, the goal is to learn the values of the weights and biases that lead to the smallest errors between actual and predicted solutions. The goal of training machine learning models is to find the values of these trainable parameters (weights for neural networks, coefficients for polynomial regression, etc). However, many machine learning models require parameters that are not learned during training. These unlearned parameters which are referred to as hyperparameters (e.g., number of layers in a neural network, or the number of trees in random forests) need to be selected by the user. The selection of these hyperparameters is important since they greatly affect the performance of the models. Selecting severely suboptimal hyperparameters will lead to poor predictive capabilities of the machine learning surrogates, and therefore, a degradation of the predicted design optimum eventually. This study employs an automated strategy to select the appropriate hyperparameters, by framing this process as another optimization problem. In the hyperparameter selection process, the objective function to be minimized is the CVMSE, while the tunable variables are the ML hyperparameters. Thus, the goal is to find the hyperparameter values that lead to the smallest CVMSE. In this study, hyperparameter tuning is performed for all the base learners independently.

The hyperparameter tuning process is performed using an open-source implantation of Bayesian Optimization (BO).[19] Bayesian optimization finds the optimum by probabilistically sampling the surface based on an acquisition function over successive iterations. The process starts by generating an initial database containing different hyperparameter settings and their corresponding CVMSE. A Gaussian process regression model is trained using this database to predict the CVMSE as a function of the hyperparameters. An important feature of BO is that it incorporates uncertainty into the sampling process, using an acquisition function that takes into account the mean and standard deviations of the CVMSE predicted by the Gaussian process regressor. The mean represents a frequentist solution, while the standard deviation gives a measure of uncertainty in the model's prediction. In this work, the acquisition function used is given by:

$$\theta = \mu + \kappa\sigma \qquad (2)$$

where $\mu$ is the mean, $\sigma$ is the standard deviation, and $\kappa$ is a constant used to control the balance between exploration and exploitation. In this study, $\kappa$ was set to a fixed value of 2.576. The acquisition function is referred to as the upper confidence bound. A global optimizer is used to minimize this function, thereby finding a point that has a favorable mean CVMSE or high model uncertainty. The actual CVMSE is evaluated at the optimum point to obtain ground truth information, and this is added to the database. This process of training the Gaussian process regressor, minimizing the acquisition function, and acquiring more ground truth data is continued until convergence. For the hyperparameter tuning process, the variables are the hyperparameters of the machine learning models, while the objective is to find the hyperparameter values that would result in the trained machine learning models having the smallest CVMSE. The BO process is performed for each base learner independently, as though it is the only model being used for prediction. Moreover, hyperparameter tuning is only performed once, using the initial simulation dataset. This hyperparameter tuning serves the sole purpose of informing how to set the untrainable hyperparameters that are needed for training the base models. Once this process is complete, the active learning loop begins. During the active learning loop, no further tuning is performed on the hyperparameters – they remain fixed throughout the loop. However, a retrained Super Learner model is obtained each time new data from CFD is added. This Super Learner model is obtained by retraining the base models individually (using the best hyperparameters for each base learner obtained using BO before the active learning loop began), and combining them using Eq. 1. It should be noted that training the Super Learner and hyperparameter tuning are different processes. Training the Super Learner involves training the individual base learners, then combining them as described in Eq. 1. This training process requires the user to pre-set the hyperparameters of the base models. Hyperparameter tuning is the process of finding the best hyperparameters to use for training the base learners that make up the Super Learner. These hyperparameters are not changed while training the base machine learning models and combining them in the Super Learner, but instead, are chosen *a priori* using BO.

In addition to automatically selecting hyperparameters, the number of samples to be included in the initial dataset for the Super Learner training needs to be determined. ML-GA involves running some initial simulations to populate a database containing the design variables and their objective values. But, the amount of data required will vary for different problems, depending on the number of design variables and the degree of non-linearity prevalent within the design space. If the number of samples generated for training is too small, the surrogate model would be a poor representation of the CFD solution. On the other hand, too many samples would be wasteful of computational resources. Therefore, in addition to automated hyperparameter tuning, active learning is also incorporated in this study to help tackle this issue. In this case, identical to ML-GA, an initial dataset consisting of $N$ samples, is generated for training the Super Learner. The optimal hyperparameters are found, the Super Learner is trained, and a global optimizer is used to find the optimum based on the trained Super Learner. However, despite techniques to improve accuracy, the global optimum obtained based on the surrogate model may produce bad objectives when evaluated using CFD. Therefore, active learning is used to refine the design objective close to the best-known design optimum. To do this, the predicted global optimum is evaluated using CFD, the new solution obtained is added to the database, and the Super Learner is retrained from scratch. Because of the new simulations, the Super Learner has more information around the global optimum and is able to make more accurate predictions in this region. This process, referred to as the active learning loop, is repeated until the convergence criteria are satisfied or when the number of maximum active learning iterations is reached. To accelerate convergence in the present work, the refinements to the global optimum involved running 5 new CFD simulations in parallel during each active learning iteration. These 5 points were obtained by finding 5 distinct global optima, obtained by repeating the Super Learner training and running the GA to find the predicted global optima. In this study, for convergence to be reached, it was required that the difference between actual and predicted objective values at the best point from the previous iteration remain below a threshold, $\varepsilon$, for four successive active learning iterations, without improvements to the maximum design objective. The implication of this is that the Super Learner accurately predicts the CFD solution at the design optimum, and thus no further improvement to the objective can be expected by continuing the loop for additional iterations. More details of the convergence criteria are presented in section 4.

## 3    Problem Setup

### 3.1    CFD model

In this study, AutoML-GA was used to perform design optimization of a turbocharged Cummins heavy-duty compression ignition engine[20] operating on a gasoline-like fuel at medium-load conditions. The engine incorporates features like exhaust gas recirculation and charged air cooling. The surrogate used for the gasoline-like fuel was a RON70 primary reference fuel (PRF) blend containing 70% iso-octane and 30% n-heptane by mass. The engine had a displacement of 14.9 *L*, a bore of 137 *mm*, and a stroke of 169 *mm*. Optimization was performed based on an engine speed of 1375 *rpm*, with 0.1245 *g* of fuel injected per cycle/cylinder over a duration of 15.58 °CA. The fuel injection temperature was 360 *K*, while the global equivalence ratio was set to 0.57. The engine operating conditions are summarized in Table 1.

Table 1. Summary of engine operating conditions [20].

| Engine model | Cummins ISX15 |
|---|---|
| Displacement | 14.9 L |
| Bore | 137 mm |
| Stroke | 169 mm |
| Connecting rod | 262 mm |
| Compression ratio | 17.3:1 |
| Engine speed | 1375 rpm |
| Intake valve closing | -137 °CA after top dead center (ATDC) |
| Exhaust valve opening | 148 °CA ATDC |
| Injection duration | 15.58 °CA |
| Mass of fuel injected | 0.1245 g/cycle/cylinder |
| Fuel injection temperature | 360 K |
| Global equivalence ratio | 0.57 |

In this study, 3D simulations were run to evaluate the objective as a function of the design variables. These simulations were run using a commercial CFD software, CONVERGE[21]. As opposed to modeling the entire cylinder, azimuthal periodicity was assumed, permitting the entire computational domain to be modeled as a sector mesh to reduce computational cost. Various models were used to capture the multi-physics phenomena involving gas dynamics, convective heat transfer, spray injection and breakup, and chemical kinetics. Turbulent flow in the cylinder was modeled using a Reynolds-averaged Navier-Stokes re-normalized group[22] model (RANS RNG $k$-$\varepsilon$). The blob injection model introduced by Reitz and Diwakar was used to initialize the fuel droplets. The Kelvin-Helmholtz Rayleigh-Taylor model[23] and the Schmidt and Rutland model[24] were used to model the spray breakup and collision between spray parcels, respectively. The evolution of species was captured using a reduced mechanism containing 48 species and 152 chemical reactions.[25] The evolution of soot was captured by the Hiroyasu soot[26] and the Nagle and Strickland-Constable[27] model, using acetylene ($C_2H_2$) as the soot precursor. The models used in the study have been validated against experiments in a previous study, where complete details regarding the CFD model are presented.[20]

In this study, five key metrics, referred to as the intermediate objectives, are extracted from the CFD simulations and used to construct the overall objective function to be maximized. These intermediate objectives are the indicated specific fuel consumption (ISFC), the maximum pressure rise rate (MPRR), the mass of soot produced ($M_{soot}$), the mass of $NO_x$ produced ($M_{NOx}$), and the maximum in-cylinder pressure (PMAX). The $M_{soot}$ and $M_{NOx}$ quantities used are normalized by the power output. Combining these output quantities, an objective function referred to as the merit, is defined as:[8]

$$F = 100 * \left\{ \frac{160}{ISFC} - 100 * f(PMAX) - 10 * f(MPRR) - f(M_{soot}) - f(M_{NOx}) \right\} \quad (2)$$

where,

$$f(PMAX) = \begin{cases} \frac{PMAX}{220} - 1, & PMAX > 220 \, bar \\ 0, & PMAX \leq 220 \, bar \end{cases} \quad (3)$$

$$f(MPRR) = \begin{cases} \frac{MPRR}{15} - 1, & MPRR > 15 \, bar/CA \\ 0, & MPRR \leq 15 \, bar/CA \end{cases} \quad (4)$$

$$f(M_{soot}) = \begin{cases} \frac{M_{soot}}{0.0268} - 1, & M_{soot} > 0.0268 \, g/kWh \\ 0, & M_{soot} \leq 0.0268 \, g/kWh \end{cases} \quad (5)$$

$$f(M_{NOx}) = \begin{cases} \frac{M_{NOx}}{1.34} - 1, & M_{NOx} > 1.34 \, g/kWh \\ 0, & M_{NOx} \leq 1.34 \, g/kWh \end{cases} \quad (6)$$

From Eqs. 2-6, it is evident that the merit function to be maximized involves minimizing ISFC while obeying certain soft constraints. The other terms, besides ISFC, do not contribute to the merit value if they stay within pre-defined constraints. However, if they cross these thresholds, the merit at that design point is penalized, in proportion to the degree to which the imposed limits are violated. Hence, by maximizing the merit, designs that reduce fuel consumption while maintaining low soot and $NO_x$ emissions are discovered. In addition, the PMAX and MPRR terms in the merit function encourages the optimizer to find designs that do not exceed the mechanical limits of the engine.

In order to achieve the objective of minimizing fuel consumption while obeying important constraints, it is necessary to adjust one or more engine design variables within pre-defined bounds. These design variables are related to fuel injector design and strategies, as well as initial in-cylinder thermodynamic and flow conditions. The list of design variables and corresponding bounds is shown in Table 2.

Table 2. List of design variables and their bounds.

| Parameter | Description | Min | Max | Units |
|---|---|---|---|---|
| nNoz | Number of nozzle holes | 8 | 10 | - |
| TNA | Total nozzle area | 1 | 1.3 | - |
| Pinj | Injection pressure | 1400 | 1800 | bar |
| SOI | Start of injection timing | -11 | -7 | °CA ATDC |
| NozzleAngle | Nozzle half-inclusion angle | 72.5 | 83.0 | degrees |
| EGR | EGR fraction | 0.35 | 0.5 | - |
| Tivc | IVC temperature | 323 | 373 | K |
| Pivc | IVC pressure | 2.0 | 2.3 | bar |
| SR | Swirl ratio | -2.4 | -1 | - |

## 3.2 AutoML-GA parameters

The AutoML-GA optimization approach used in this study starts by generating a set of initial design points using Latin Hypercube Sampling with Multi-Dimensional Uniformity,[28, 29] and running CFD simulations for the generated design points. After performing automated hyperparameter tuning, this initial database is used to train the various base learners in the Super Learner, as described in section 2. In section 3.1, the merit, made up of five intermediate objectives, was described. As opposed to training the Super Learner to predict the merit directly, five separate Super Learners were trained for each intermediate objective to improve accuracy. The Super Learner-predicted merit was then obtained in the same manner as the ground truth by using Eqs. 2-6. A list of all the hyperparameters, their default values, and their bounds within which they were tuned, is shown in Table 3. The Super Learner was trained using an in-house code developed in Python 3.6, using packages such as NumPy[17] and Scikit-learn.[18] The default hyperparameter values were based on the default settings in Scikit-learn version 0.22.2.

Table 3. List of hyperparameters with the corresponding default values and bounds.

| Base learner | Parameter | Default values | Min value | Max value | Log scale |
|---|---|---|---|---|---|
| RPR | Regularization weight, $\alpha$ | 0 | -6 | 0 | Yes |
| RPR | Polynomial degree | 2 | 1 | 10 | No |
| SVR | Parameter to control the number of support vectors, $\upsilon$ | 0.5 | $1\times10^{-10}$ | 1.0 | No |
| SVR | penalty parameter, $C$ | 0 | -6 | 2.5 | Yes |
| SVR | Kernel coefficient, $\gamma$ | Inverse of the number of features | -6 | 0 | Yes |
| KRR | Regularization parameter, $\alpha$ | 0 | -6 | 0 | Yes |
| KRR | Gamma parameter, $\gamma$ | None | -4 | 0 | Yes |
| XGB | Number of trees | 3 | 2 | 4 | Yes |

| XGB | Learning rate | -1 | -3 | 0 | Yes |
| XGB | Maximum tree depth | 6 | 2 | 8 | No |
| ANN | Number of neurons in each hidden layer | 100 | 10 | 250 | No |
| ANN | L2 regularization parameter, $\alpha$ | -4 | -6 | 0 | Yes |
| ANN | Tolerance to determine if training should be stopped early, *tol* | -4 | -6 | -2 | Yes |

As mentioned in section 2, an active learning loop that involves running more simulations at the Super Learner-predicted optima and adding these points to the database successively was incorporated. Each active learning iteration involved running five CFD simulations, obtained by repeating the Super Learner training five times and using a global optimizer each time to get a predicted optimum. Since some base learners (e.g., the neural network, XGBoost) have non-deterministic components, the Super Learner obtained and the resulting predicted global optimum shows some degree of variation each time this is performed. Therefore, five unique CFD points were added to the existing database of points after every active learning iteration.

## 4   Results

In this section, results comparing the performance of AutoML-GA (which involves automated hyperparameter tuning and active learning) and ML-GA (default hyperparameters with active learning) are compared. After each iteration, the difference between the actual and predicted optimum, ε, was computed to assess convergence. In this study, since five simulations were run per active learning iteration, only the best-performing point was used to calculate this difference. Convergence was assumed to be achieved when ε remains below $1\times10^{-3}$ without an improvement in the maximum merit for four active learning iterations. Since a single test of AutoML-GA and ML-GA may cause misleading conclusions, ten trials were performed for both to provide a more statistically meaningful result. However, for all trials, the active learning loop was run for a fixed number of 20 iterations (including the simulations for data generation) even if they converged earlier, so that the average performance could be plotted as a function of iteration number. Furthermore, both ML-GA and AutoML-GA were tested with two initial sample sizes, $N = 150$ and $N = 200$. For AutoML-GA, a time limit of 4 hours was imposed on the hyperparameter tuning process. Each CFD simulation was run on a single node comprising 36 Intel Xeon E5-2695 processors. For AutoML-GA, it took a runtime of 86.5 hours to tune the hyperparameters and complete the active learning iteration. In comparison, ML-GA took 82.6 hours to run the active learning loop. Each active learning loop, on an average, had a runtime of 4.3 hours, so performing automated hyperparameter tuning was roughly equivalent to running an extra active learning iteration.

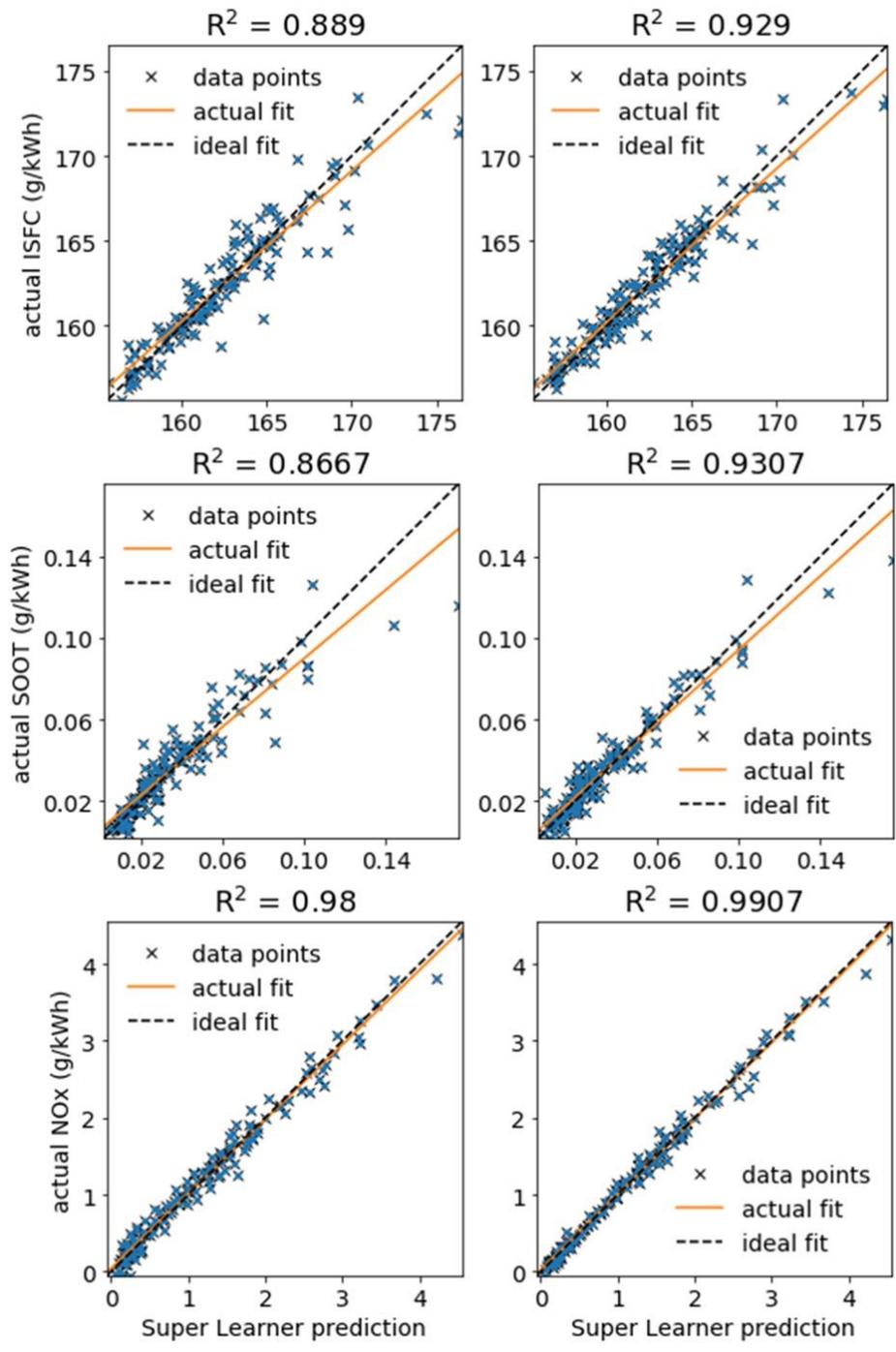

Figure 3. Super Learner parity plots obtained using ML-GA (left) and AutoML-GA (right).

Table 4. $R^2$ values of ML-GA and AutoML-GA for various learners and targets.

| | ANN | | | | |
|---|---|---|---|---|---|
| | ISFC | $M_{soot}$ | $M_{NOx}$ | MPRR | PMAX |
| ML-GA | 0.8784 | 0.851 | 0.975 | 0.929 | 0.976 |
| AutoML-GA | 0.915 | 0.895 | 0.9897 | 0.9443 | 0.9907 |
| | SVR | | | | |
| | ISFC | $M_{soot}$ | $M_{NOx}$ | MPRR | PMAX |
| ML-GA | 0.8306 | 0.7896 | 0.918 | 0.925 | 0.973 |
| AutoML-GA | 0.9033 | 0.8975 | 0.984 | 0.947 | 0.993 |
| | XGB | | | | |
| | ISFC | $M_{soot}$ | $M_{NOx}$ | MPRR | PMAX |
| ML-GA | 0.5977 | 0.637 | 0.8257 | 0.667 | 0.6724 |
| AutoML-GA | 0.809 | 0.7837 | 0.922 | 0.81 | 0.884 |
| | KRR | | | | |
| | ISFC | $M_{soot}$ | $M_{NOx}$ | MPRR | PMAX |
| ML-GA | 0.763 | 0.6963 | 0.863 | 0.8706 | 0.955 |
| AutoML-GA | 0.911 | 0.9077 | 0.984 | 0.9443 | 0.993 |
| | RPR | | | | |
| | ISFC | $M_{soot}$ | $M_{NOx}$ | MPRR | PMAX |
| ML-GA | 0.8467 | 0.814 | 0.9097 | 0.9106 | 0.95 |
| AutoML-GA | 0.9214 | 0.924 | 0.982 | 0.947 | 0.9893 |
| | Super Learner | | | | |
| | ISFC | $M_{soot}$ | $M_{NOx}$ | MPRR | PMAX |
| ML-GA | 0.8867 | 0.8623 | 0.9756 | 0.938 | 0.979 |
| AutoML-GA | 0.9316 | 0.934 | 0.9907 | 0.952 | 0.994 |

A comparison of $R^2$ values of the predicted values in relation to the actual quantities are shown in Table 4. The $R^2$ values are a measure of how well the values predicted by the machine learning model match the ground truth data. While the Super Learner is retrained before each active learning iteration, it is impractical to show the $R^2$ values for all iterations. Instead, here the results shown here represent the predictions before the active learning loop begins, i.e., based on the initial dataset. Moreover, the results are based on the initial sample size of $N = 150$. Lower error levels, which are desirable, lead to $R^2$ values that are close to 1, while less accurate predictions by the machine learning models lead to lower $R^2$ values. From Table 4, it can be seen that for all cases, the Super Learner leads to $R^2$ values that are better than the best-performing base learner. Secondly, it can be seen that for all the individual base learners and the Super Learner, AutoML-GA produces predictions with higher $R^2$ values, compared to ML-GA. This is true for all targets, showing that the default hyperparameters are suboptimal and that the optimized hyperparameters lead to much better predictions. In addition to Table 4, Fig. 3 shows scatter plots of the actual values of ISFC, $M_{soot}$, and $M_{NOx}$, against the values predicted by the Super Learner, for both ML-GA and AutoML-GA. Ideally, all the scatter points would lie on the dashed diagonal line that shows the ideal fit. In practice, inevitable errors lead to an imperfect fit, shown by the solid diagonal line in the plots. Overall, it can be observed that the actual fit obtained using AutoML-GA is closer to the ideal fit, compared to the ML-GA fit, with less scatter of points from the ideal diagonal line.

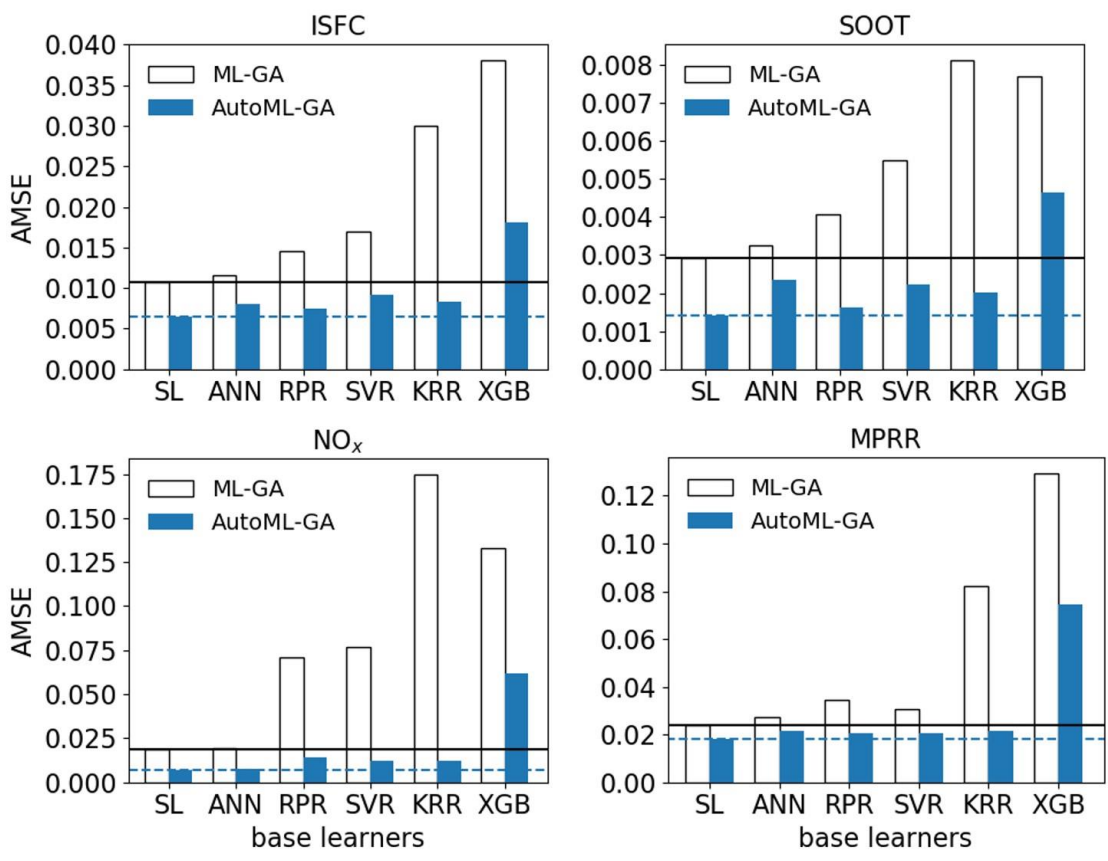

Figure 4. Mean squared errors averaged over multiple trials for various learners and target variables ($N = 150$). The hollow bars and colored bars represent the ML-GA and AutoML-GA results, respectively. The solid and dashed horizontal lines indicate the Super Learner AMSE for ML-GA and AutoML-GA, respectively.

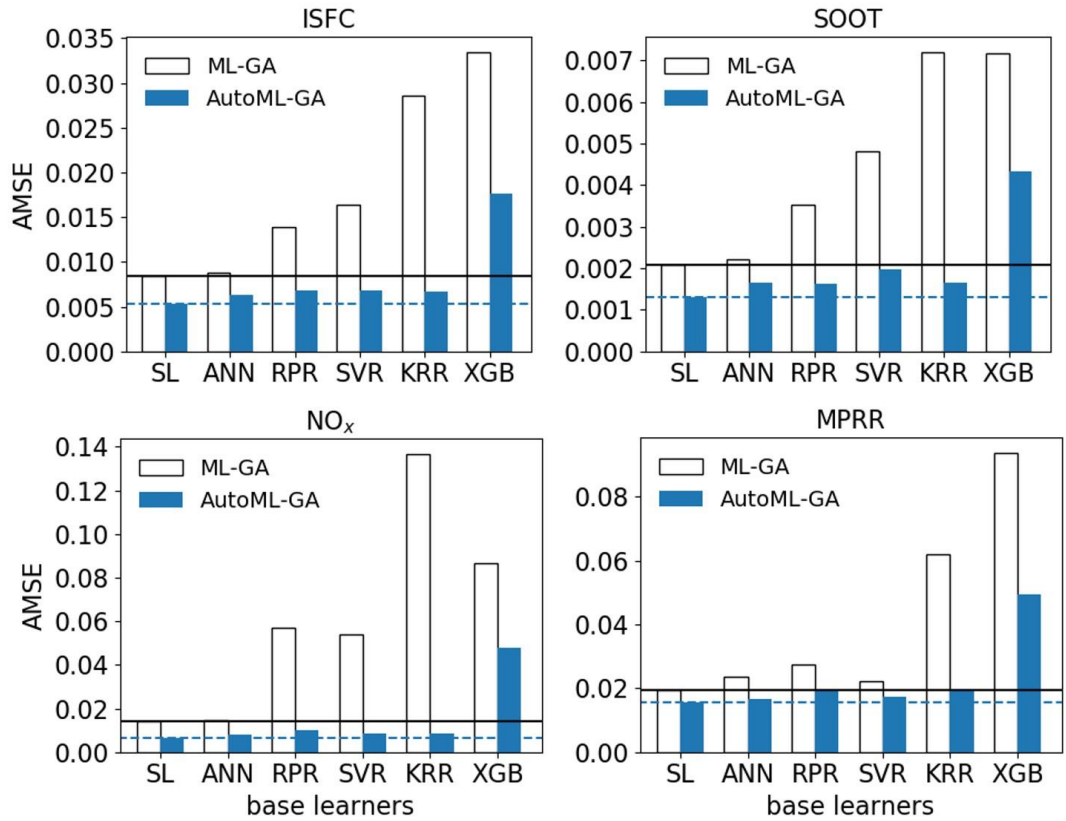

Figure 5. Mean squared errors averaged over multiple trials for various learners and target variables ($N = 200$). The hollow bars and colored bars represent the ML-GA and AutoML-GA results, respectively. The solid and dashed horizontal lines indicate the Super Learner AMSE for ML-GA and AutoML-GA, respectively.

The prediction errors of ML-GA compared with those of AutoML-GA are presented in Figs. 4 and 5, where mean squared errors, averaged over all the ten trials (AMSE), are shown. Figures 4 and 5 are based on initial sample size, $N = 150$ and 200, respectively. The ML-GA and AutoML-GA Super Learner AMSE are represented by solid and dashed horizontal lines respectively, to provide a visual comparison with the base learners. In Fig. 4 ($N = 150$), it can be seen that for all the intermediate objectives shown, the Super Learner outperforms all the base learners. It can also be observed that AutoML-GA leads to much lower prediction errors compared to ML-GA. The degree of reduction in error varies for different targets and base learners. In particular, the benefits of using AutoML-GA are most evident for PMAX, where the prediction of the base learners sees a 74% reduction in the AMSE on average, and least for MPRR, with 42%. The Super Learner AMSE sees 39%, 52%, 62%, 24%, and 72% decrease for ISFC, $M_{soot}$, $M_{NOx}$, MPRR, and PMAX, respectively. Overall, Fig. 4 shows that surrogate models can provide much better representation of the CFD model when automated hyperparameter tuning is performed. An examination of Fig. 5 shows that similar conclusions can be reached for the case with $N = 200$. Similar AMSE reductions of 40% and 73% are obtained for MPRR and PMAX, respectively. In general, ML-GA with $N = 200$ shows significantly lower errors compared to the case with $N = 150$, suggesting that the model accuracy is still significantly increasing with the data sample size. In this case, the Super Learner AMSE decreased by 26%, 40%, 33%, 25%, and 34% for ISFC, $M_{soot}$, $M_{NOx}$, MPRR, and PMAX, respectively. This is because the errors are still much larger than the irreducible error, and thus there is potential for improvement with the inclusion of more data samples. However, for AutoML-GA, the errors show less difference with 20%, 7.6%, 9.7, 19%, and 6.4% for ISFC, $M_{soot}$, $M_{NOx}$, MPRR, and PMAX, respectively. This suggests that for most intermediate objectives, AutoML-GA is much closer to the irreducible error already, and adding more samples will only lead to modest improvements in accuracy. In Figs. 4 and 5, it can be seen that for ML-GA, the Super Learner test error is similar to the errors obtained from the artificial neural network, suggesting that the other base learners (especially KRR and XGB) barely contribute to the stacked Super Learner model. On the other hand, For AutoML-GA, KRR is comparable to the other base learners, while XGB's errors remain high. While XGB could have been removed from the list of base learners for this problem, excluding it as a base learner may not be of much value because the hyperparameter optimization process is performed in parallel. Thus, by removing base learners, computing resources may be conserved (since the number of models to train is lower) but the overall runtime would still be similar. Keeping a large list of base learners is also useful since the relative performance of the base learners is problem-dependent. For example, for a different design problem, XGB may be more important than say, RPR.

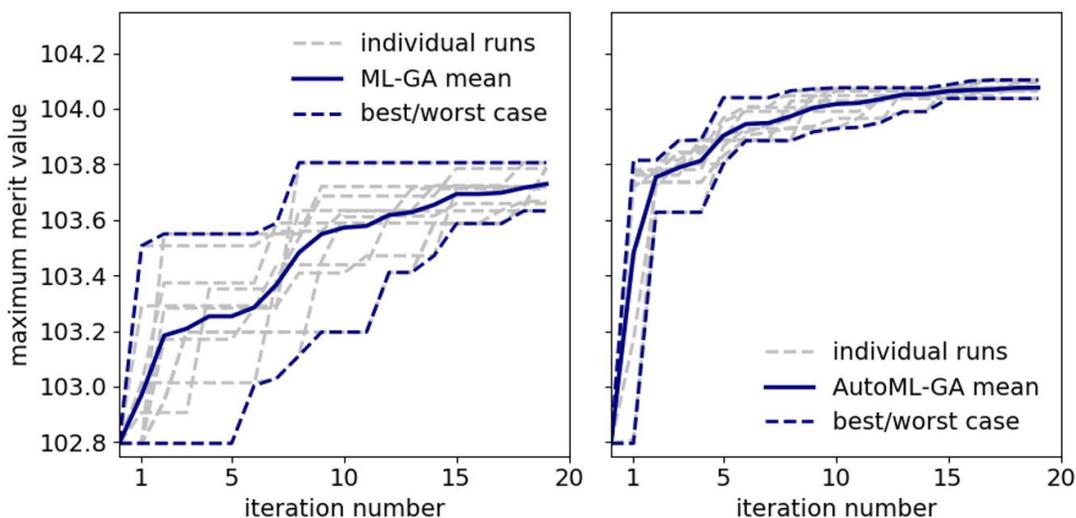

Figure 6. Maximum merit values as a function of the number of active learning iterations using ML-GA (left) and AutoML-GA (right). The plots are based on $N = 150$.

Figure 6 shows the maximum merit as a function of the number of active learning iterations in ML-GA and AutoML-GA for $N = 150$. In the figure, ML-GA is shown on the left, while AutoML-GA is shown on the right. The grey lines show the individual trials, the dashed lines represent the best and worst cases, and the solid lines represent the maximum merit averaged over all trials. Two observations are evident from Fig. 6. First, it can be seen that AutoML-GA produces better designs compared to ML-GA. The merit improves by $6.7 \times 10^{-3}$ for AutoML-GA and $1.7 \times 10^{-3}$ for ML-GA on average after the first iteration, and the difference between the two approaches only grows as the number of iterations increases. It takes 2, 5 and 9, iterations for the mean merit values of AutoML-GA to reach a merit of 103.5, 103.85, and 104.0, respectively. ML-GA is comparatively less efficient, taking 9 iterations to reach 103.5. ML-GA does not reach 103.85, with a maximum merit of 103.73 at the end of the active learning loop. Most striking, however, is that the worst case for AutoML-GA outperforms the best case using ML-GA, with a merit of 104.0 after 20 iterations for AutoML-GA's worst case and 103.8 for ML-GA's best case.

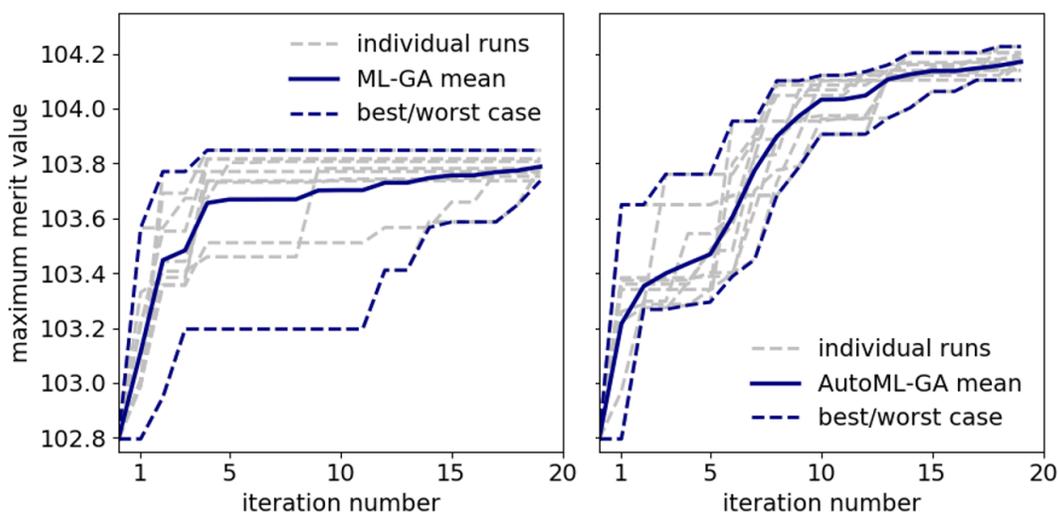

Figure 7. Maximum merit values as a function of the number of active learning iterations using ML-GA (left) and AutoML-GA (right). The plots are based on $N = 200$.

Figure 7 shows the same information as Fig. 7 except that the plots are for $N = 200$. Here again, it is seen that AutoML-GA converges faster than ML-GA. Despite having an outlying bad trial (the worst case), the mean maximum merit produced by ML-GA is higher for $N = 200$ compared to $N = 150$. On the other hand, AutoML-GA using $N = 150$ appears to perform better than $N = 200$ during the initial stages before falling behind as the 20th iteration is approached. At the end of 20 iterations, AutoML-GA with $N = 200$ achieves a higher mean merit of 104.17 (compared with $N = 150$ that reaches 104.08). The better performance of $N = 150$ during the earlier iterations, while counterintuitive, is perhaps due to statistical noise, as a result of the small number of trials that are performed.

Table 5. Baseline design and best designs obtained from ML-GA and AutoML-GA.

|  | Baseline[8] | ML-GA | AutoML-GA |
|---|---|---|---|
| **Design variables** | | | |
| **nNoz** | 9 | 10 | 10 |
| **SOI (°CA ATDC)** | -9.00 | -9.00 | -10.09 |
| **TNA** | 1.00 | 1.02 | 1.00 |
| **NozzleAngle (degrees)** | 152.00 | 155.02 | 158.31 |
| **SR** | -1.00 | -2.40 | -1.68 |
| **EGR** | 0.41 | 0.44 | 0.44 |
| **Pinj (bar)** | 1600.00 | 1794.15 | 1416.41 |
| **Pinv (bar)** | 215000 | 229623 | 229688 |
| **Tivc (K)** | 323.00 | 324.33 | 324.32 |
| **Output variables** | | | |
|  | Baseline | ML-GA | AutoML-GA |
| **ISFC (g/kWh)** | 156.53 | 154.13 | 153.69 |
| $M_{SOOT}$ **(g/kWh)** | 0.0235 | 0.010345 | 0.014007 |
| $M_{NOx}$ **(g/kWh)** | 1.07 | 1.32 | 1.31 |
| **MPRR (bar/CA)** | 11.22 | 14.07 | 10.86 |
| **PMAX (bar)** | 152.31 | 161.85 | 159.93 |
| **MERIT** | 102.2 | 103.81 | 104.10 |

Table 5 shows a breakdown of the optimum designs obtained by ML-GA and AutoML-GA and compares them with the Baseline design. The upper section of the Table shows the design variables, while the lower section shows the various target outputs contained in the merit function. The best cases correspond to merit values of 103.81 and 104.10 for ML-GA and AutoML-GA, respectively. From Table 5, it can be seen that for both the ML-GA and AutoML-GA optimum designs, values of $M_{SOOT}$, $M_{NOx}$, MPRR, and PMAX, remain below the thresholds defined in Eqs. 3-6. Therefore, these constraints do not have any effect on the merit values at the design optimum, since the optimizer is able to find designs that cause the negative contributions to the right-hand side of Eq. 2 to be 0. The table also shows that the ISFCs are 156.53 g/kWh, 154.13 g/kWh, and 153.69 g/kWh, for the Baseline, ML-GA, and AutoML-GA, respectively. Therefore, by using ML-GA, the fuel consumption reduces by 1.53% compared to the Baseline, while it reduces by 1.81% for AutoML-GA. Figure 8 shows the temporal evolution of the in-cylinder pressure for the three cases. The plots show consistency with the ISFC values in Table 5. Since the mass of injected fuel is constant, designs with lower ISFC are expected to produce more work per cycle. As expected, AutoML-GA and ML-GA have much higher in-cylinder pressures compared to the Baseline.

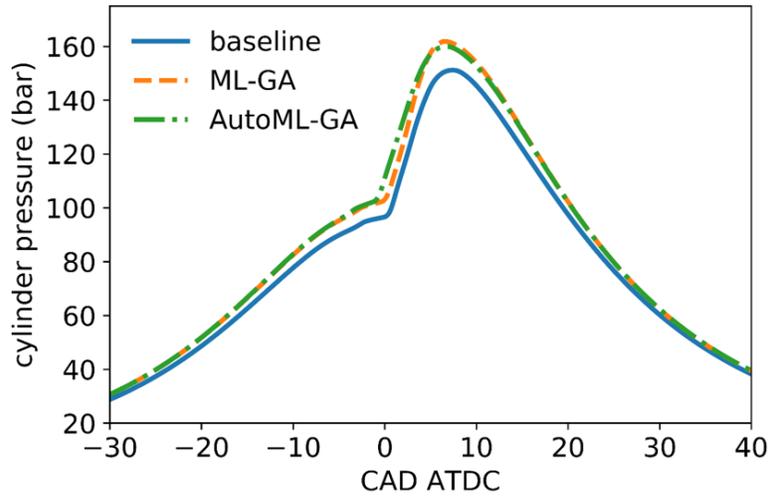

Figure 8. Comparison of the temporal in-cylinder pressure evolution for the Baseline, ML-GA best case, and AutoML-GA best case.

To better understand the improved merits achieved by ML-GA and AutoML-GA's optimum designs, CFD contour plots (on a vertical cut plane) of equivalence ratio and temperature are shown in Fig. 9 at 15 crank angle degrees after the top dead center (ATDC). The plots compare the Baseline case with the best cases from ML-GA and AutoML-GA. The temperature plots (left), show that ML-GA and AutoML-GA lead to higher in-cylinder temperatures compared to the baseline. This is consistent with the output variables for the cases in Table 5, where it can be seen that ML-GA and AutoML-GA lead to lower soot, and slightly higher NOx and peak pressures. On the plots showing the equivalence ratio (right), it can be seen that much larger isolated pockets of rich fuel-air mixtures exist in the Baseline case, compared to the best designs from ML-GA and AutoML-GA. This is likely due to better fuel-air mixing as a result of the larger number of injector holes and/or higher swirl ratio. While the ML-GA and AutoML-GA best designs show similar EGR and TNA in Table 5, other design variables are different. For instance, ML-GA optimum maintains the same SOI timing as the Baseline, while AutoML-GA selects an earlier injection timing. This earlier start of injection perhaps results in better mixing, as evident from the smaller pockets of rich fuel-air mixture for AutoML-GA compared to ML-GA. Interestingly, the injection pressure for the AutoML-GA optimum is lower than the Baseline and ML-GA optimum, which would also indicate lower cost of the injection system.

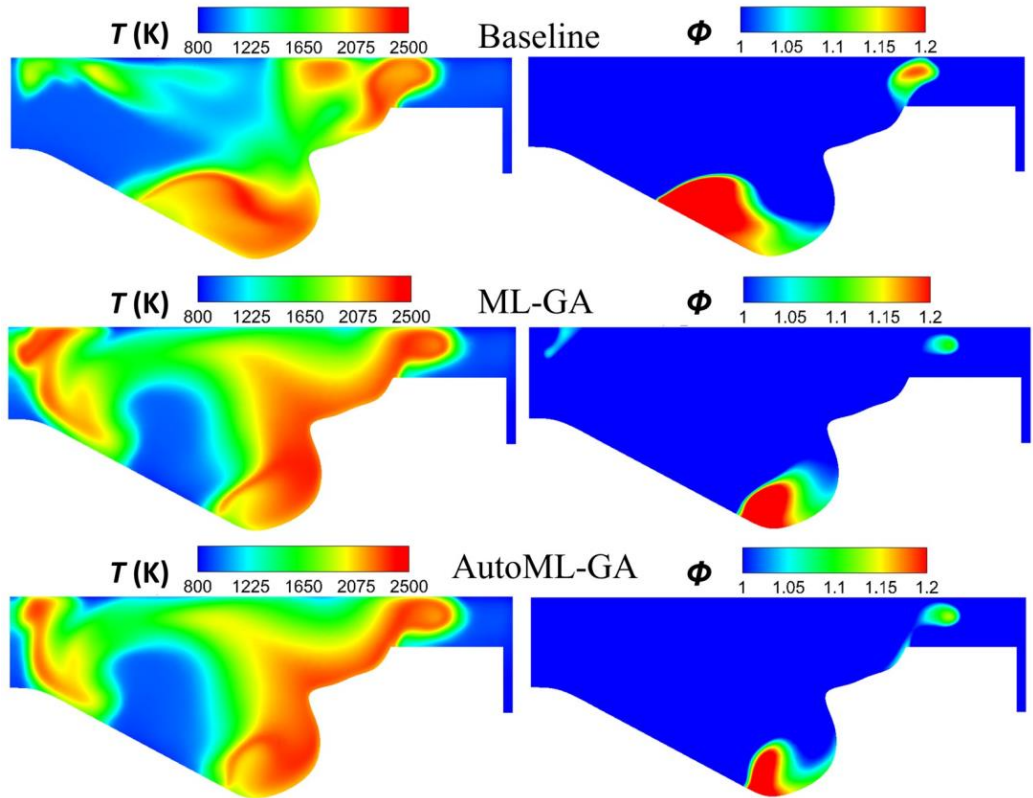

Figure 9. Equivalence ratio and temperature contour plots for the Baseline case (top), ML-GA best case (middle), and AutoML-GA best case (bottom).

The evolution of ISFC and NOx emissions with the number of active learning iterations is shown in Fig. 10 for AutoML-GA. The plots are population-based figures for $N = 150$, where each iteration has 5 circle symbols, each corresponding to the 5 design optima (as discussed in section 2.2 and illustrated in Fig. 2). The plots show that the ISFC starts at a high value on average, and decreases as the optimization progresses. On the contrary, the amount of NOx produced increases, but most points remain below the threshold. The merits at points that cross the threshold (depicted by the dashed line in the plot) are penalized. Ultimately, the design optimum obtained by AutoML-GA is one where $M_{NOx}$ remains just below the limit while having the lowest value of ISFC.

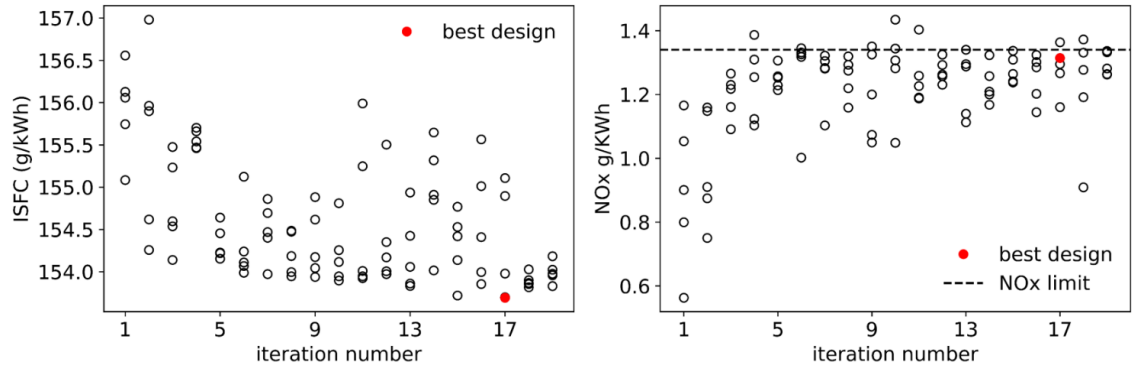

Figure 10. Population-based plots showing the evolution of ISFC (left) and $M_{\text{NOx}}$ (right) with the number of active learning iterations. The plots are based on $N = 150$.

As mentioned in section 2, the behavior of ε, which is the difference between the predicted and actual merit at each iteration, is used to assess convergence. In Fig. 11, plots showing epsilon against the iteration number is shown. It can be observed that for ML-GA (left), epsilon shows minimal decay and remains at relatively high values, compared to AutoML-GA (right), where epsilon decays to zero more readily as more iterations are performed. This shows that even after several iterations, the optimum predicted by ML-GA varies significantly from what is obtained when the results are validated in CFD. On the contrary, in the vicinity of the best-known point, AutoML-GA is able to provide predictions that are close to the actual CFD solutions after just a few iterations.

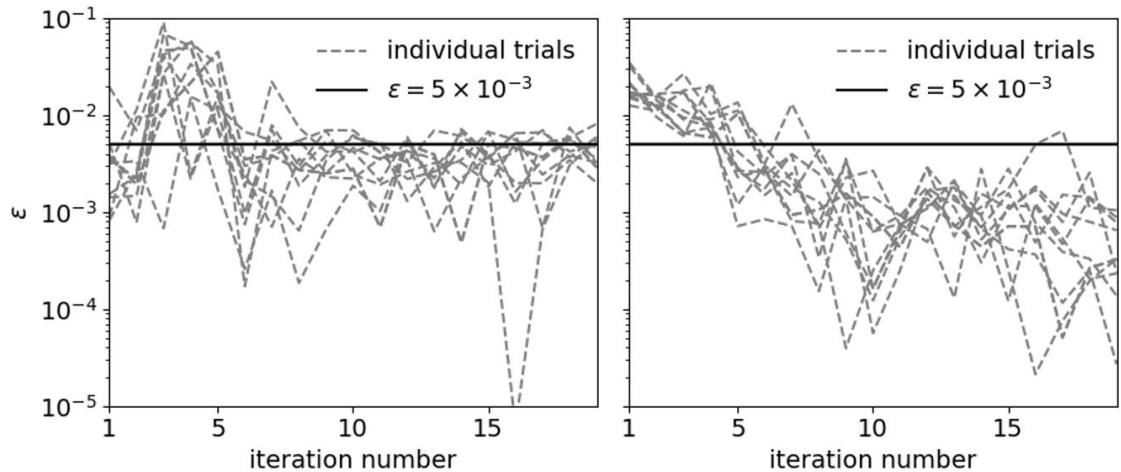

Figure 11. The evolution of ε using ML-GA (left) and AutoML-GA (right). The plots are based on $N = 150$.

Therefore, as expected, AutoML-GA is able to reach convergence within 20 active learning iterations for all cases. On average, it takes 10 and 9.8 iterations for AutoML-GA to reach convergence with $N = 150$ and $N = 200$, respectively. Since the values of $\varepsilon$ remain high for ML-GA, the optimizer never reaches a state of convergence for any of the trials run in this study.

Table 6. List of optimal hyperparameters.

| | ISFC | $M_{soot}$ | $M_{NOx}$ | MPRR | PMAX |
|---|---|---|---|---|---|
| **RPR** | | | | | |
| Regularization weight, $\alpha$ | 3 | 3 | 3 | 3 | 3 |
| Polynomial degree | $1.66 \times 10^{-2}$ | $5.6 \times 10^{-4}$ | $2.02 \times 10^{-2}$ | $4.24 \times 10^{-2}$ | $2.2 \times 10^{-2}$ |
| **XGB** | | | | | |
| | ISFC | $M_{soot}$ | $M_{NOx}$ | MPRR | PMAX |
| Maximum tree depth | 2 | 2 | 2 | 2 | 2 |
| Learning rate | 0.0948 | 0.1360 | 0.0904 | 0.0498 | 0.0733 |
| Number of trees | 2620 | 2728 | 4354 | 7155 | 3351 |
| **SVR** | | | | | |
| | ISFC | $M_{soot}$ | $M_{NOx}$ | MPRR | PMAX |
| Kernel coefficient, $\gamma$ | 0.2041 | 0.1246 | 0.0658 | 0.0644 | 0.0465 |
| Parameter to control the number of support vectors, $\upsilon$ | 0.5411 | 0.6001 | 0.8391 | 0.6671 | 0.8665 |
| Penalty parameter, $C$ | 144.7 | 253.6 | 153.8 | 86.9 | 239.7 |
| **ANN** | | | | | |
| | ISFC | $M_{soot}$ | $M_{NOx}$ | MPRR | PMAX |
| Number of neurons in each hidden layer | 183 | 196 | 188 | 194 | 177 |
| L2 regularization parameter, $\alpha$ | $2.83 \times 10^{-6}$ | $2.01 \times 10^{-6}$ | $7.98 \times 10^{-6}$ | $2.19 \times 10^{-6}$ | $3.77 \times 10^{-6}$ |

| Tolerance to determine if training should be stopped early, *tol* | $9.97 \times 10^{-4}$ | $2.36 \times 10^{-3}$ | $1.23 \times 10^{-2}$ | $3.07 \times 10^{-3}$ | $1.7 \times 10^{-3}$ |
|---|---|---|---|---|---|
| **KRR** | | | | | |
|  | ISFC | $M_{soot}$ | $M_{NOx}$ | MPRR | PMAX |
| Gamma parameter, $\gamma$ | 0.0383 | 0.0329 | 0.0511 | 0.0729 | 0.0216 |
| Regularization parameter, $\alpha$ | $1.01 \times 10^{-5}$ | $2.06 \times 10^{-6}$ | $2.12 \times 10^{-5}$ | $3.72 \times 10^{-4}$ | $2.37 \times 10^{-6}$ |

Finally, Table 6 shows the optimal hyperparameters obtained using automated hyperparameter tuning. These hyperparameters are based on $N = 150$, and tuning is performed using the initial dataset before the active learning loop begins. From the table, it can be observed that while there are some exceptions, in most cases the optimal hyperparameters depend on the target variable of interest. For instance, for the RPR model, the polynomial degree for all targets is 3. However, the degree of regularization, which is meant to reduce overfitting varies widely. For instance, $M_{soot}$ requires much lower regularization compared to MPRR, suggesting that predicting the mass of soot requires the model to be more flexible compared to the pressure rise rate. These trends are not evident by observing the raw data. Manually finding unique hyperparameters for each of these targets by trial-and-error, even with the aid of domain-knowledge, would be a daunting task. Secondly, a comparison with Table 3 shows that oftentimes, the optimized hyperparameters of AutoML-GA are different from the default values used in ML-GA. An example of this can be seen for XGB, where the default number of trees, being 1000, appears to be too small for this problem. For all targets, AutoML-GA employs a much larger number of trees. Using a larger number of trees, as evident in Figs. 3 and 4, assists in lowering the prediction errors. This suggests, that with a relatively small dataset consisting of 150 samples, using 1000 trees leads to high variance. Overall, automated adjustment of hyperparameters in AutoML-GA leads to better surrogate models, and as a consequence, better merit values from optimization. While 150 and 200 samples were tested in this study, hyperparameter selection would likely be important for data sizes within the range of practical data sizes for engine design optimization ($O(100) - O(1000)$ data points). This is because issues regarding the selection of hyperparameters, including the avoidance of overfitting and underfitting, would remain relevant concerns to handle.

While this study involved coupling AutoML-GA with simulations for CFD-driven optimization, the end goal of this process is the design of more efficient and cleaner engines. In practice, the design optimum obtained from the AutoML-GA needs to be verified by building physical prototypes and running engine tests. In this regard, the development of accurate computational models is essential to obtaining good agreement between the predicted performance improvement based on the CFD-driven design optimization and actual benefits observed from experiments. On the other hand, AutoML-GA can also be directly coupled with experiments to perform engine calibration.

## 5  Conclusion

In this study, an automated approach to building ensemble surrogate models for design optimization, AutoML-GA, was presented. By incorporating an automated approach to selecting hyperparameters, the proposed approach reduces the need for prior machine learning expertise and the tedious trial-and-error process of tuning hyperparameters. In addition to the automation of hyperparameter tuning, an active learning loop was incorporated, where the design optimum discovered from the surrogate model was further refined by iteratively running more CFD simulations. In this paper, the advantages of AutoML-GA was demonstrated by applying it to the optimization of a compression-ignition IC engine operating on gasoline-like fuel, to find the design variables that reduce fuel consumption while obeying pre-defined constraints associated with emissions and engine mechanical limits. First, it was shown that by incorporating automated hyperparameter tuning, the surrogate models produced much better representations of the ground truth simulations, compared to the use of default hyperparameters. Furthermore, it was demonstrated that better surrogate models translated to better merit values. In the case where the size of the initial dataset ($N$) was 150, on average AutoML-GA converged after a total of 10 iterations to a merit of 104.0. On the other hand, even after running for 20 iterations, ML-GA did not attain convergence, and only reached a mean merit value of 103.73 and 103.79, for $N = 150$ and $N = 200$, respectively. Analysis of the selected hyperparameters showed that the optimal hyperparameters were very different from the default hyperparameters, and varied depending on the target variable of interest. Future studies will focus on demonstrating the proposed approach for engine design optimization considering multiple speed-load points and piston bowl design.

## 6  Funding

The submitted manuscript has been created by UChicago Argonne LLC, Operator of Argonne National Laboratory (Argonne). The U.S. Government retains for itself, and others acting on its behalf, a paid-up nonexclusive, irrevocable world-wide license in said article to reproduce, prepare derivative works, distribute copies to the public, and perform publicly and display publicly, by or on behalf of the Government. This work was supported by the U.S. Department of Energy, Office of Science under contract DE-AC02-06CH11357. The research work was funded by the Department of Energy Technology Commercialization Fund project TCF-18-15349 and DOE SBIR grants DE-SC0020464 and DE-SC0019695.

## 7   Acknowledgments

The authors acknowledge the computing resources available through "Blues," a high-performance computing cluster operated by the Laboratory Computing Resource Center (LCRC) at Argonne National Laboratory.

14. Močkus J. On Bayesian methods for seeking the extremum. In: *Optimization techniques IFIP technical conference* 1975, pp.400-404. Springer.

15. Drucker H, Burges CJ, Kaufman L, et al. Support vector regression machines. In: *Advances in neural information processing systems* 1997, pp.155-161.

16. Chen T and Guestrin C. Xgboost: A scalable tree boosting system. In: *Proceedings of the 22nd acm sigkdd international conference on knowledge discovery and data mining* 2016, pp.785-794.

17. Walt Svd, Colbert SC, Varoquaux GJCiS, et al. The NumPy array: a structure for efficient numerical computation. 2011; 13: 22-30.

18. Pedregosa F, Varoquaux G, Gramfort A, et al. Scikit-learn: Machine learning in Python. *Journal of Machine Learning Research* 2011; 12: 2825-2830.

19. Nogueira F. Bayesian Optimization: Open source constrained global optimization tool for Python. 2014.

20. Pal P, Probst D, Pei Y, et al. Numerical investigation of a gasoline-like fuel in a heavy-duty compression ignition engine using global sensitivity analysis. *SAE International Journal of Fuels and Lubricants* 2017; 10: 56-68.

21. Richards K, Senecal P and Pomraning E. CONVERGE 2.3. *Convergent Science, Madison, WI* 2019.

22. Han Z, Reitz RD and technology. Turbulence modeling of internal combustion engines using RNG κ-ε models. *Combustion Science and Technology* 1995; 106: 267-295.

23. Reitz RD. Modeling atomization processes in high-pressure vaporizing sprays. *Atomisation and Spray Technology* 1987; 3: 309-337.

24. Schmidt DP and Rutland C. A new droplet collision algorithm. *Journal of Computational Physics* 2000; 164: 62-80.

25. Liu Y-D, Jia M, Xie M-Z, et al. Enhancement on a skeletal kinetic model for primary reference fuel oxidation by using a semidecoupling methodology. *Energy Fuels* 2012; 26: 7069-7083.

26. Hiroyasu H and Kadota T. Models for combustion and formation of nitric oxide and soot in direct injection diesel engines. *SAE transactions* 1976: 513-526.

27. Nagle J and Strickland-Constable RF. Oxidation of Carbon between 1000-2000°C. In: *Proceedings of the 5th Conference on Carbon* 1962, Pergamon Press.

28. Deutsch JL and Deutsch CV. Latin hypercube sampling with multidimensional uniformity. *Journal of Statistical Planning* 2012; 142: 763-772.

29. Moza S. sahilm89/lhsmdu: Latin Hypercube Sampling with Multi-Dimensional Uniformity (LHSMDU): Speed Boost minor compatibility fixes. 1.1.1 ed.: Zenodo, 2020.
28